\documentclass[10pt,letterpaper,twocolumn]{article}
\usepackage[latin9]{inputenc}
\usepackage{color}
\usepackage{array}
\usepackage{float}
\usepackage{booktabs}
\usepackage{multirow}
\usepackage{amsmath}
\usepackage{amssymb}
\usepackage{graphicx}
\usepackage{mathtools}
\usepackage{subcaption}
\usepackage{threeparttable}
\usepackage[breaklinks=true,bookmarks=false]{hyperref}
\usepackage{hyperref}
\hypersetup{
    colorlinks=true,
    linkcolor=red,
    urlcolor=red,
}
\makeatletter
\newcommand{\mypar}[1]{\smallskip\noindent {\bf #1}\enskip}

\providecommand{\tabularnewline}{\\}
\floatstyle{ruled}
\newfloat{algorithm}{tbp}{loa}
\providecommand{\algorithmname}{Algorithm}
\floatname{algorithm}{\protect\algorithmname}

\usepackage{cvpr}
\usepackage{times}
\usepackage{epsfig}
\usepackage{graphicx}
\usepackage{url}
\cvprfinalcopy 

\def\cvprPaperID{6422} 

\makeatother
\begin{document}
\title{Instance Credibility Inference for Few-Shot Learning}
\author{Yikai Wang\textsuperscript{1,4}\qquad 
Chengming Xu\textsuperscript{1}\qquad 
Chen Liu\textsuperscript{1}\qquad 
Li Zhang\textsuperscript{2}\qquad 
Yanwei Fu\textsuperscript{1,3,4}\thanks{Corresponding author.}\\
\textsuperscript{1}School of Data Science, Fudan University\\
\textsuperscript{2}Department of Engineering Science, University of Oxford\\
\textsuperscript{3}MOE Frontiers Center for Brain Science, Fudan University\\
\textsuperscript{4}Shanghai Key Lab of
Intelligent Information Processing, Fudan University\\
{\tt\small 
\{yikaiwang19, cmxu18, chenliu18, yanweifu\}@fudan.edu.cn,
lz@robots.ox.ac.uk
}
}

\maketitle
\begin{abstract}
Few-shot learning (FSL) aims to recognize new objects with extremely limited training data for each category. 
Previous efforts are made by either leveraging meta-learning paradigm or novel principles in data augmentation to alleviate this extremely data-scarce problem. 
In contrast, 
this paper presents a simple statistical approach, 
dubbed Instance Credibility Inference (ICI) to exploit the distribution support of unlabeled instances for few-shot learning. 
Specifically, 
we first train a linear classifier with the labeled few-shot examples and use it to infer the pseudo-labels for the unlabeled data. 
To measure the credibility of each pseudo-labeled instance, 
we then propose to solve another linear regression hypothesis by increasing the sparsity of the incidental parameters and rank the pseudo-labeled instances with their sparsity degree. 
We select the most trustworthy pseudo-labeled instances alongside the labeled examples to re-train the linear classifier. 
This process is iterated until all the unlabeled samples are included in the expanded training set, 
i.e. the pseudo-label is converged for unlabeled data pool.
Extensive experiments under two few-shot settings show that our simple approach can establish new state-of-the-arts on four widely used few-shot learning benchmark datasets including \textit{mini}ImageNet, \textit{tiered}ImageNet, CIFAR-FS, and CUB. 
Our code is available at: \url{https://github.com/Yikai-Wang/ICI-FSL}
\end{abstract}

\section{Introduction}
\begin{figure*}[hbt!]
\begin{centering}
\includegraphics[width=2\columnwidth]{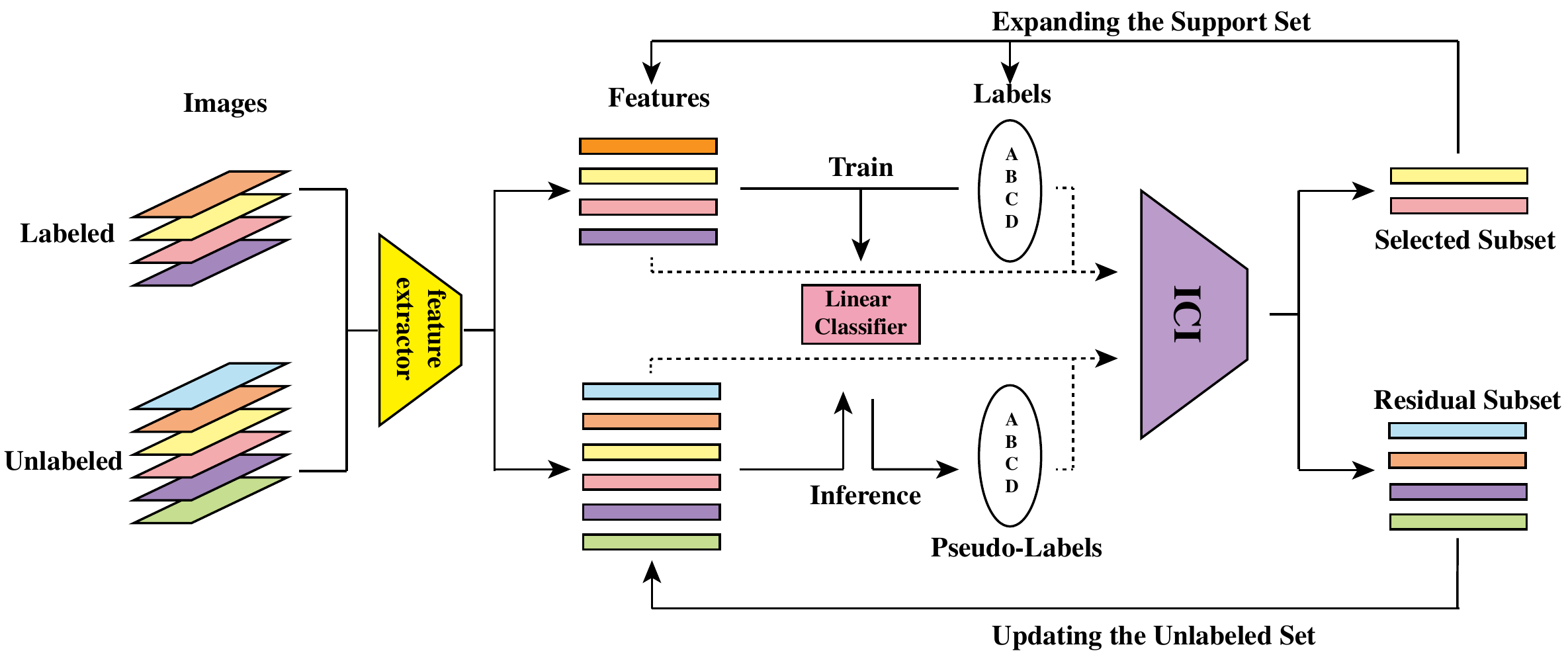}
\par\end{centering}
\caption{\label{fig:framework} 
Schematic illustration of our proposed framework. 
In the inference process of $N$-way-$m$-shot FSL task with unlabeled data, 
we embed each instance, 
inference each unlabeled data and use ICI to select the most trustworthy subset to expand the support set. 
This process is repeated until all unlabeled data are included in the support set.}
\end{figure*}

Learning from one or few examples is an important ability for humans.
For example, children have no problem forming the concept of ``giraffe'' by only taking a glance from a picture in a book, 
or hearing its description as looking like a deer with a long neck~\cite{zhang2017learning}.
In contrast, the most successful recognition systems~\cite{krizhevsky2012imagenet,simonyan2014very,he2016deep,huang2017densely} still highly rely on an avalanche of labeled training data.
This thus increases the burden in rare data collection (\eg accident data in the autonomous driving scenario) and
expensive data annotation (\eg disease data for medical diagnose), 
and more fundamentally limits their scalability to open-ended learning of the long tail categories in the real-world.

Motivated by these observations, there has been a recent resurgence of research interest in few-shot learning~\cite{finn2017model,snell2017prototypical,sung2018learning,vinyals2016matching}.
It aims to recognize new objects with extremely limited training data for each category. 
Basically, a few-shot learning model has the chance to access the source/base dataset with many labeled training instances for model training and then is able to generalize to a disjoint but relevant target/novel 
dataset
with only scarce labeled data.
A simplest baseline to transfer learned knowledge to the novel set is fine-tuning~\cite{yosinski2014transferable}.
However, it would cause severely overfitting as one or a few instances are insufficient to model the data distributions of the novel classes.
Data augmentation and regularization techniques can alleviate overfitting in such a limited-data regime, but they do not solve it.
Several recent efforts are made in leveraging learning to learn, or meta-learning paradigm by simulating the few-shot scenario in the training process~\cite{lemke2015metalearning}. 
However, Chen \etal~\cite{DBLP:journals/corr/abs-1904-04232} empirically argue that such a learning paradigm often results in inferior performance compared to a simple baseline with a linear classifier coupled with a deep feature extractor.

Given such a limited-data regime (one or few labeled examples per category), one of the fundamental problems for few-shot learning is that one can hardly estimate the data distribution without introducing the inductive bias.
To address this problem, 
two types of strategy resort to model the data distribution of novel category beyond traditional \emph{inductive} few-shot learning:
(i) semi-supervised few-shot learning (SSFSL)~\cite{liu2018learning,ren2018meta,sun2019learning} supposes that we can utilize 
unlabeled data (about ten times more than labeled data) to help to learn the 
model;
furthermore,
(ii) transductive inference~\cite{joachims1999transductive} for few-shot learning (TFSL)~\cite{liu2018learning,qiao2019transductive} assumes we can access to all the test data, rather than evaluate them one by one in the inference process. 
In other words, the few-shot learning model can utilize the data distributions of testing examples. 

Self-taught learning~\cite{self-taught-learning} is one of the most straightforward ways in leveraging the information of unlabeled data. Typically, a trained classifier infers the labels of unlabeled data, which are further taken to update the classifier. 
Nevertheless, the inferred pseudo-labels may not be always trustworthy; the wrongly labeled instances may jeopardize the performance of the classifier.
It is thus essential to investigate the labeling confidence of each unlabeled instance.

To this end, we present a simple statistical approach, dubbed Instance Credibility Inference (ICI) to exploit the distribution support of unlabeled instances for few-shot learning. 
Specifically, we first train a linear classifier (\eg, logistic regression) with the labeled few-shot examples and use it to infer the pseudo-labels for the unlabeled data.
Our model aims to iteratively select the most trustworthy pseudo-labeled instances according to their credibility measured by the proposed ICI to augment the training set. 
The classifier thus can be progressively updated and further infer the unlabeled data.
We iterate this process until all the unlabeled samples are included in the expanded training set, \ie the pseudo-label is converged for unlabeled data pool.
The schematic illustration is shown in Figure~\ref{fig:framework}. 

Basically, we re-purpose the standard self-taught learning algorithm by our ICI algorithm.
How to select the pseudo-labeled data to exclude the wrong-predicted samples,~\ie, excluding the noise introduced by the self-taught learning strategy?
Our intuition is that the algorithm of sample selection can neither rely only on the label space (\eg based on the probability of each class given by the classifier) nor the feature space (\eg select samples most similar to training data). 
Instead, 
we introduce a linear regression hypothesis by regressing each instance (labeled and pseudo-labeled) from feature to label space and increase the sparsity of the incidental parameter~\cite{fan2012partial} until it vanishes.
Thus we can rank pseudo-labeled instances with sparsity degree as their credibility.
We conduct extensive experiments on major few-shot learning datasets to validate the effectiveness of our proposed algorithm. 

The contributions of this work are as follows:
(i) 
We present a simple statistical approach, dubbed Instance Credibility Inference (ICI) to exploit the distribution support of unlabeled instances for few-shot learning. 
Specifically, our model iteratively selects the pseudo-labeled instances according to its credibility measured by the proposed ICI for classifier training.
(ii) 
We re-purpose the standard self-taught learning algorithm~\cite{self-taught-learning} by our proposed ICI. To measure the credibility of each pseudo-labeled instance, we solve another linear regression hypothesis by increasing the sparsity of the incidental parameter~\cite{fan2012partial} and rank the sparsity degree as the credibility for each pseudo-labeled instance.
(iii)
Extensive experiments under two few-shot settings show that our simple approach can establish new state-of-the-arts on four widely used few-shot learning benchmark datasets including \textit{mini}ImageNet, \textit{tiered}ImageNet, CIFAR-FS, and CUB.
\section{Related work}
\mypar{Semi-supervised learning.} 
Semi-supervised learning (SSL) aims to improve the learning performance with limited labeled data by exploiting large amount of unlabeled data.
Conventional approaches focus on finding the low-density separator within both labeled and unlabeled data~\cite{vapnik1998statistical,bennett1999semi,joachims1999transductive},
and avoid to learn the ``wrong'' knowledge from the unlabeled data~\cite{li2014towards}.
Recently, 
semi-supervised learning with deep learning models use consistency regularization~\cite{conf/iclr/LaineA17},
moving average technique~\cite{tarvainen2017mean} and
adversarial perturbation regularization~\cite{miayto2016virtual}
to train the model with large amount of unlabeled data.
The key difference between semi-supervised learning and few-shot learning with unlabeled data is that the unlabeled data is still limited in the latter.
To some extent, the low-density assumption widely utilized in SSL is hard to achieve in the few-shot scenario, making SSFSL a more difficult problem.

Self-taught learning~\cite{self-taught-learning}, also known as self-training~\cite{NoisyStudent},
is a traditional semi-supervised strategy of utilizing unlabeled data to improve the performance of classifiers~\cite{amini2002semi,grandvalet2005semi}. 
Typically, an initially trained classifier predicts class labels of unlabeled instances; the  unlabeled data with  pseudo-labels are further selected to update the classifier.~\cite{lee2013pseudo}. 
Current algorithms based on self-taught learning includes training neural networks using labeled data and pseudo-labeled data jointly~\cite{lee2013pseudo}, 
using mix-up between unlabeled data and labeled data to reduce the influence of noise~\cite{arazo2019pseudo}, 
using label propagation for pseudo-labeling based on a nearest-neighbor graph and measuring the credibility using entropy~\cite{iscen2019label},
and re-weighting the pseudo-labeled data based on the cluster assumption on the feature space~\cite{shi2018transductive}. 
Unfortunately, the predicted pseudo-labels may not be trustworthy.
Different and orthogonal to previous re-weighting or mix-up works,  we design a statistical algorithm in estimating the credibility of each instance assigned with its corresponding pseudo-label. Only the most confident instances are employed to update the classifier.

\mypar{Few-shot learning.}
Recent efforts on FSL are made towards the following aspects.
(1) Metric learning methods, putting emphasis on finding better distance metrics, 
include weighted nearest neighbor classifier (\eg Matching Network~\cite{vinyals2016matching}), 
finding prototype for each class (\eg Prototypical Network~\cite{snell2017prototypical}),
or learning specific metric for each task (\eg TADAM~\cite{oreshkin2018tadam});
(2) Meta learning methods, such as Meta-Critic~\cite{sung2017learning}, MAML~\cite{finn2017model}, Meta-SGD~\cite{li2017meta}, Reptile~\cite{nichol2018first}, and LEO~\cite{rusu2018meta},  optimize the models for the capacity of rapidly adapted to new tasks. 
(3) Data augmentation algorithms enlarge available data to alleviate the lack of data in the image level~\cite{chen2019image} or the feature level~\cite{ren2018meta}. Additional, SNAIL~\cite{mishra2017simple} utilizes the sequence modeling to create a new framework.
The proposed statistical algorithm is orthogonal but potentially useful to improve these algorithms -- it is always worth increasing the training set by utilizing the unlabeled data with confidently predicted labels.

\mypar{Few-shot learning with unlabeled data.}
Recently approaches  tackle few-shot learning problems by resorting to additional  unlabeled data. Specifically,  in semi-supervised few-shot learning settings, recent works~\cite{ren2018meta,liu2018learning} enables unlabeled data from the same categories to better handle the true distribution of each class. Furthermore, transductive settings have also been considered recently. For example,  LST~\cite{sun2019learning} utilizes self-taught learning strategy in a meta-learning manner. 
Different from these methods, this paper presents a conceptually simple statistical approach derived from self-taught learning; our approach, empirically and significantly improves the performance of FSL on several benchmark datasets, by only using very simple classifiers, \eg, logistic regression, or Support Vector Machine (SVM).
\section{Methodology}

\subsection{Problem formulation}

We introduce the formulation of few-shot learning here. Assume a base category
set $\mathcal{C}_{base}$, and a novel category set $\mathcal{C}_{novel}$
with $\mathcal{C}_{base}\bigcap\mathcal{C}_{novel}=\emptyset$. 
Accordingly, the base and novel datasets are $D_{base}=\left\{ \left(\mathbf{I}_{i},y_{i}\right),y_{i}\in\mathcal{C}_{base}\right\} $,
and $D_{novel}=\left\{ \left(\mathbf{I}_{i},y_{i}\right),y_{i}\in\mathcal{C}_{novel}\right\} $,
respectively. 
In few-shot learning, the recognition models on $D_{base}$
should be generalized to the novel category $C_{novel}$ with only one
or few training examples per class. 

For evaluation, we adopt the standard \emph{$N$-way-$m$-shot} classification as \cite{vinyals2016matching} on $D_{novel}$. 
Specifically, in each episode, we randomly sample $N$ classes $L\sim C_{novel}$;
and $m$ and $q$ labeled images per class are randomly sampled in $L$ to construct the support set $S$ and the query set $Q$, respectively.
Thus we have $\left|S\right|=N\times m$ and $\left|Q\right|=N\times q$.
The classification accuracy is averaged on query sets $Q$ of many
meta-testing episodes. In addition, we have unlabeled
data of novel categories $U_{novel}=\left\{ \mathbf{I}_{u}\right\} $. 

\subsection{Self-taught learning from unlabeled data}

In general, labeled data for machine learning is often very difficult
and expensive to obtain, while the unlabeled data can be utilized
for improving the performance of supervised learning. Thus we recap
the self-taught learning formalism -- one of the most classical semi-supervised
methods for few-shot learning~\cite{self-taught-learning}. Particularly,
assume $f\left(\cdot\right)$ is the feature extractor trained
on the base dataset $D_{base}$. One can train a supervised classifier
$g\left(\cdot\right)$ on the support set $S$, and pseudo-labeling unlabeled data, $\hat{y}_{i}=g\left(f\left(\mathbf{I}_{u}\right)\right)$
with corresponding confidence $p_{i}$ given by the classifier. The most confident unlabeled
instances will be further taken as additional data of corresponding
classes in the support set $S$. Thus we obtain the updated supervised
classifier $\hat{g}\left(\cdot\right)$. To this end, few-shot classifier
acquires additional training instances, and thus its performance can
be improved. 

However, it is problematic if directly utilizing self-taught learning
in one-shot cases. Particularly, the supervised classifier $g\left(\cdot\right)$
is only trained by few instances. The unlabeled instances with high
confidence may  not be correctly categorized, and the classifier will
be updated by some wrong instances. Even worse, one can not assume
the unlabeled instances follows the same class labels or generative
distribution as the labeled data. Noisy instances or outliers may
also be utilized to update the classifiers. To this end, we propose
a systematical algorithm: Instance Credibility Inference (ICI) to reduce the noise.

\subsection{Instance credibility inference (ICI)}
To measure the credibility of predicted labels over unlabeled data, we introduce a hypothesis of linear model by regressing each instance from feature to label spaces.  Particularly, given $n$  instances of $N$ classes,
$S=\left\{ \left(\mathbf{I}_{i},y_{i},\mathbf{x}_{i}\right),y_{i}\in\mathcal{C}_{novel}\right\} $, where $y_i$ is the ground truth when $\mathbf{I}_{i}$ come from the support set, or the pseudo-label when $\mathbf{I}_{i}$ come from the unlabeled set, we employ a simple linear regression model to ``predict'' the class label,
\begin{equation}
y_{i}=\mathbf{x}_{i}^{\top}\beta+\gamma_{i}+\epsilon_{i}\label{eq:lm},
\end{equation}

\noindent where $\beta\in\mathcal{\mathbb{R}}^{d\times N}$ is the
coefficient matrix for classification; $\mathbf{x}_{i}\in\mathcal{\mathbb{R}}^{d\times1}$
is the feature vector of instance $i$; $y_{i}$ is $N$ dimension
one-hot vector denoting the class label of instance $i$. 
Note that to facilitate the computations, we employ PCA~\cite{tipping1999probabilistic} to reduce the dimension of extracted features $f\left(\mathbf{I}_{i}\right)$ to $d$. $\epsilon_{ij}\sim\mathcal{N}\left(0,\sigma^{2}\right)$
is the Gaussian noise of zero mean and $\sigma$ variance. 
Inspired by incidental parameters~\cite{fan2012partial}, 
we introduce $\gamma_{i,j}$ to amend the chance of instance $i$ belonging to class  $y_{j}$.
Larger $\left\Vert\gamma_{i,j}\right\Vert$, the higher difficulty in attributing instance $i$ to  class  $y_{j}$.

\begin{figure}
\begin{centering}
\includegraphics[width=1\columnwidth]{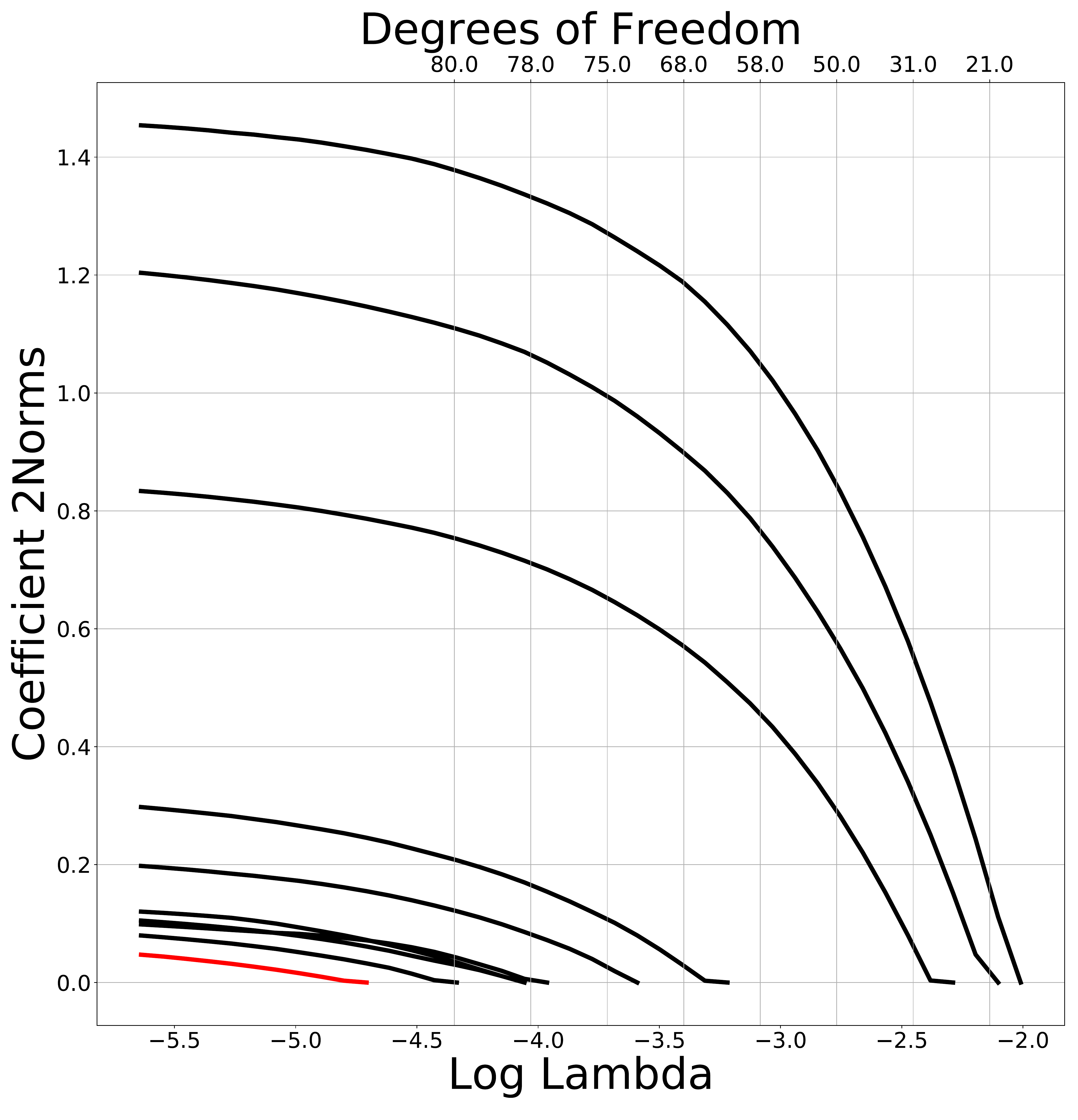}
\par\end{centering}
\caption{\label{fig:illu}
Regularization path of $\lambda$ on ten samples. Red line is corresponding to the most trustworthy sample suggested by our ICI algorithm.}
\end{figure}

Write Eq.~\ref{eq:lm} in a matrix form for all instances, we are thus solving the problem of:
\begin{equation}
\left(\hat{\beta},\hat{\gamma}\right)=\underset{\beta,\gamma}{\arg\min}\left\Vert Y-X\beta-\gamma\right\Vert _{F}^{2}+\lambda R\left(\gamma\right)\label{eq:loss_func},
\end{equation}
where $\left\Vert \cdot \right\Vert _{F}^{2}$ denotes the Frobenius norm. $Y=\left[y_{i}\right]\in\mathcal{\mathbb{R}}^{n\times N}$ and
$X=\left[\mathbf{x}_{i}^{\top}\right]\in\mathcal{\mathbb{R}}^{n\times d}$
indicate label and feature input respectively. $\gamma=\left[\gamma_{i}\right]\in\mathcal{\mathbb{R}}^{n\times N}$
is the incidental matrix, with the penalty $R\left(\gamma\right)=\sum_{i=1}^{n}\left\Vert \gamma_{i}\right\Vert _{2}$.
$\lambda$ is the coefficient of penalty. To solve Eq.~\ref{eq:loss_func},
we re-write the function as 
\[
L\left(\beta,\gamma\right)=\left\Vert Y-X\beta-\gamma\right\Vert _{F}^{2}+\lambda R\left(\gamma\right).
\]
Let $\frac{\partial L}{\partial\beta}=0$, we have 
\begin{equation}
\hat{\beta}=\left(X^{\top}X\right)^{\dagger}X^{\top}\left(Y-\gamma\right)\label{eq:beta},
\end{equation}

\noindent where $\left(\cdot\right)^{\dagger}$ denotes the Moore-Penrose pseudo-inverse. 
Note that 
(1) we are interested in utilizing $\gamma$
to measure the credibility of each instance along its regularization path, rather than estimating
$\hat{\beta}$, since the linear regression model is not good enough
for classification in general. 
(2) the $\hat{\beta}$ also relies
on the estimation of $\gamma$. 
To this end, we take Eq.~\ref{eq:beta}
into $L\left(\cdot\right)$ and solve the problem
as,
\begin{equation}
\underset{\gamma\in\mathbb{R}^{n\times N}}{\arg\min}\left\Vert Y-H\left(Y-\gamma\right)-\gamma\right\Vert _{F}^{2}+\lambda R\left(\gamma\right),
\end{equation}
where 
$H=X\left(X^{\top}X\right)^{\dagger}X^{\top}$
is the hat matrix of
$X$.
We further define $\tilde{X}=\left(I-H\right)$ and $\tilde{Y}=\tilde{X}Y$.
Then the above equation can be simplified as
\begin{equation}
\underset{\gamma\in\mathbb{R}^{n\times N}}{\arg\min}\left\Vert \tilde{Y}-\tilde{X}\gamma\right\Vert _{F}^{2}+\lambda R\left(\gamma\right),\label{eq:penalty}
\end{equation}
which is a multi-response regression problem. 
We seek the best subset by checking the regularization path, 
which can be easily configured by 
a blockwise descent algorithm
implemented in Glmnet~\cite{simon2013blockwise}. 
Specifically,
we have a theoretical value of $\lambda_{max}=\underset{i}{\max}\left\Vert\tilde{X}_{\cdot i}^{\top}\tilde{Y}\right\Vert _{2}/n$~\cite{simon2013blockwise} to guarantee the solution of Eq.~\ref{eq:penalty} all 0.
Then we can get a list of $\lambda$s from $0$ to $\lambda_{max}$. 
We solve a specific Eq.~\ref{eq:penalty} with each $\lambda$,
and get the regularization path of $\gamma$ along the way.
Particularly, we regard $\gamma$ as a function of $\lambda$. 
When $\lambda$ changes from $0$ to $\infty$, the sparsity of $\gamma$ is increased until all of its elements are forced to be vanished. 
Further, our penalty $R\left(\gamma\right)$ encourages $\gamma$ vanishes row by row, \ie, instance by instance. 
Moreover, the penalty will tend to vanish the subset of $\tilde{X}$ with the lowest deviations, indicating less discrepancy between the prediction and the ground truth.
Hence we could rank the pseudo-labeled data by their $\lambda$ value when the corresponding $\gamma_i$ vanishes. 
As shown in one toy example of Figure~\ref{fig:illu}, the $\gamma$ value of the instance denoted by the red line vanishes first, and thus it is the most trustworthy sample by our algorithm. 
\begin{algorithm}
\textbf{Input}:support data$\left\{ \left(X_{i},y_{i}\right)\right\} _{i=1}^{N\times K}$, query data $X_{t}=\left\{ X_{j}\right\} _{j=1}^{M}$, unlabeled data $X_{u}=\left\{ X_{k}\right\} _{k=1}^{U}$

\textbf{Initialization}: support set $\left(X_{s},y_{s}\right)=\left\{ \left(X_{i},y_{i}\right)\right\} _{i=1}^{N\times K}$, feature matrix $X_{N\times K+U,d}=\left[X_{s};X_{u}\right]$, classifier

\textbf{Repeat:}

Train classifier using $\left(X_{s},y_{s}\right)$;

Get pseudo-label $y_u$ for $X_u$ by classifier;

Rank $\left(X,y\right)=\left(X,[y_s;y_u]\right)$ by ICI;

Select a subset $\left(X_{sub},y_{sub}\right)$ into $\left(X_{s},y_{s}\right)$;

\textbf{Until Converged.}

\textbf{Inference:}

Train classifier using $\left(X_{s},y_{s}\right)$;

Get pseudo-label $y_t$ for $X_t$ by classifier;

\textbf{Output}: inference labels $y_{t}=\left\{ \hat{y}_{j}\right\} _{j=1}^{M}$

\caption{\label{alg:Inference-process.}Inference process of our algorithm.}
\end{algorithm}
\subsection{Self-taught learning with ICI}

The proposed ICI can thus be easily integrated to improve the self-taught learning algorithm. Particularly, the initialized classifier can predict the pseudo-labels of unlabeled instances; and we further employ the ICI algorithm to select the most confident subset of unlabeled instances, to update the classifier. The whole algorithm can be iteratively updated, as summarized in Algorithm~\ref{alg:Inference-process.}.

\section{Experiments}

\mypar{Datasets.}
Our experiments are conducted on several widely few-shot learning benchmark datasets for general object recognition and fine-grained classification, including
\emph{mini}ImageNet~\cite{ravi2016optimization}, \emph{tiered}ImageNet~\cite{ren2018meta}, 
CIFAR-FS~\cite{bertinetto2018metalearning} and 
CUB~\cite{wah2011caltech}.
\textbf{\emph{mini}}\textbf{ImageNet} consists of $100$ classes with $600$ labeled instances in each category.
We follow the split proposed by~\cite{ravi2016optimization}, using $64$ classes as the base set to train the feature extractor, $16$ classes as the validation set and 
report performance on the novel set which consists of $20$ classes. 
\textbf{\emph{tiered}}\textbf{ImageNet} is a larger dataset compared with \emph{mini}ImageNet, and its categories are selected with hierarchical structure to split base and novel datasets semantically. 
We follow the split introduced in~\cite{ren2018meta} with base set of $20$ superclasses ($351$ classes), validation set of $6$ superclasses ($97$ classes) and novel set of $8$ superclasses ($160$ classes). 
Each class contains $1281$ images on average. 
\textbf{CUB} is a fine-grained dataset of $200$ bird categories with $11788$ images in total. 
Following the previous setting in~\cite{hilliard2018few}, we use $100$ classes as the base set, $50$ for validation and $50$ as the novel set. 
To make a fair comparison, we crop all images with the bounding box provided by ~\cite{triantafillou2017few}.
\textbf{CIFAR-FS} is a dataset with lower-resolution images derived from CIFAR-100~\cite{krizhevsky2009learning} .
It contains $100$ classes with $600$ instances in each class. 
We follow the split given by~\cite{bertinetto2018metalearning}, using $64$ classes to construct the base set, $16$ for validation and $20$ as the novel set.

\vspace{0.1in}
\mypar{Experimental setup.}
Unless otherwise specified, we use the following settings and implementation  in the experiments for our approach to make a fair comparison. 
As in~\cite{mishra2017simple,oreshkin2018tadam,lee2019meta}, we use ResNet-12~\cite{DBLP:journals/corr/HeZRS15} with $4$ residual blocks as the feature extractor in our experiments. 
Each block consists of three $3\times3$ convolutional layers, each of which followed by a BatchNorm layer and a LeakyReLu(0.1) activation. In the end of each block, a $2\times2$ max-pooling layer is utilized to reduce the output size.
The number of filters in each block is $64$, $128$, $256$ and $512$ respectively.
Specifically, referring to~\cite{lee2019meta}, we adopt the Dropout~\cite{JMLR:v15:srivastava14a} in the first two block to vanish $10\%$ of the output,
and adopt DropBlock~\cite{ghiasi2018dropblock} in the latter two blocks to vanish $10\%$ of output in channel level.
Finally, an average-pooling layer is employed to produce the input feature embedding.
We select 90\% images from each training class (\eg, 64 categories for \textit{mini}ImageNet) to construct our training set for training the feature extractor and use the remaining 10\% as the validation set to select the best model.
We use SGD with momentum as the optimizer to train the feature extractor from scratch.
Momentum factor and $L_{2}$ weight decay is set to $0.9$ and $1e-4$, respectively.
All inputs are resized to $84\times84$.
We set the initial learning rate of $0.1$, decayed by $10$ after every $30$ epochs.
The total training epochs is $120$ epochs. 
In all of our experiments, we normalize the feature with $L_2$ norm and reduce the feature dimension to $d=5$ using PCA~\cite{tipping1999probabilistic}.
Our model and all baselines are evaluated over $600$ episodes with $15$ test samples from each class.

\begin{table*}
\centering
\begin{tabular*}{\textwidth}{@{\extracolsep{\fill}}llllllllll}
\toprule 
\multirow{2}{*}{Setting} & \multirow{2}{*}{Model} &  \multicolumn{2}{c}{\emph{mini}ImageNet} & \multicolumn{2}{c}{\emph{tiered}ImageNet} & \multicolumn{2}{c}{CIFAR-FS} & \multicolumn{2}{c}{CUB}\tabularnewline
 & &$1$shot & $5$shot & $1$shot & $5$shot & $1$shot & $5$shot & $1$shot & $5$shot\tabularnewline
\midrule
\multirow{11}{*}{In.}
&Baseline$^{*}$~\cite{DBLP:journals/corr/abs-1904-04232} & $51.75$ & $74.27$ & - & - & - & - & $65.51$ & $82.85$\tabularnewline
& Baseline++$^{*}$~\cite{DBLP:journals/corr/abs-1904-04232}&$51.87$ &$75.68$ & - & - & - & - & $67.02$ & $83.58$\tabularnewline
& MatchingNet$^{*}$~\cite{vinyals2016matching}& $52.91^{\textcolor{red}{1}}$ & $68.88^{\textcolor{red}{1}}$ & - & - & - & - & $72.36^{\textcolor{red}{1}}$ & $83.64^{\textcolor{red}{1}}$\tabularnewline
& ProtoNet$^{*}$~\cite{snell2017prototypical}& $54.16^{\textcolor{red}{1}}$ & $73.68^{\textcolor{red}{1}}$ & - & - & $72.20^{\textcolor{red}{3}}$ & $83.50^{\textcolor{red}{3}}$ & $71.88^{\textcolor{red}{1}}$ & $87.42^{\textcolor{red}{1}}$\tabularnewline
& MAML$^{*}$~\cite{finn2017model}& $49.61^{\textcolor{red}{1}}$& $65.72^{\textcolor{red}{1}}$ & - & - & - & - & $69.96^{\textcolor{red}{1}}$ & $82.70^{\textcolor{red}{1}}$\tabularnewline
&RelationNet$^{*}$~\cite{sung2018learning} & $52.48^{\textcolor{red}{1}}$ & $69.83^{\textcolor{red}{1}}$ & - & - & - & - & $67.59^{\textcolor{red}{1}}$ & $82.75^{\textcolor{red}{1}}$\tabularnewline
& adaResNet~\cite{munkhdalai2017rapid}& $56.88$ & $71.94$ & - & - & - & - & - & -\tabularnewline
& TapNet~\cite{yoon2019tapnet} & $61.65$ & $76.36$ & $63.08$& $80.26$  & - & - & - & -\tabularnewline
& CTM$^{\dag}$~\cite{li2019finding} & $64.12$ & $80.51$ & $68.41$ & $84.28$  & - & - & - & -\tabularnewline
&MetaOptNet~\cite{lee2019meta}&$64.09$&$80.00$&$65.81$&$81.75$&$72.60$&$84.30$&-&-\tabularnewline
\midrule

\multirow{2}{*}{Tran.}
&TPN~\cite{liu2018learning} & $59.46$ & $75.65$ & $58.68^{\textcolor{red}{4}}$ & $74.26^{\textcolor{red}{4}}$ & $65.89^{\textcolor{red}{4}}$ & $79.38^{\textcolor{red}{4}}$ & - & -\tabularnewline
&TEAM$^{*}$~\cite{qiao2019transductive}  & $60.07$ & $75.90$ & - & - & $70.43$ & $81.25$ & $80.16$ & $87.17$ \tabularnewline
\midrule
\multirow{4}{*}{Semi.}
&MSkM with MTL & $62.10^{\textcolor{red}{2}}$ & $73.60^{\textcolor{red}{2}}$ & $68.6^{\textcolor{red}{2}}$ & $81.00^{\textcolor{red}{2}}$  & - & - & - &- \tabularnewline
&TPN with MTL & $62.70^{\textcolor{red}{2}}$ & $74.20^{\textcolor{red}{2}}$ & $72.10^{\textcolor{red}{2}}$ & $83.30^{\textcolor{red}{2}}$ & - & - & - & -\tabularnewline
&MSkM~\cite{ren2018meta}&$50.40$ & $64.40$ & $52.40$ & $69.90$ & - & - & - &  - \tabularnewline
&TPN~\cite{liu2018learning}& $52.78$& $66.42$ & $55.70$ & $71.00$ & - & - & - & -  \tabularnewline
&LST~\cite{sun2019learning}& $70.10$& $78.70$ & $77.70$ & $85.20$ & - & - & - & -  \tabularnewline
\midrule 
\midrule 
In. &LR & $56.06$ & $75.70$ & $69.02$ & $85.37$ & $62.25$ & $80.82$ & $76.16$ & $90.32$\tabularnewline
In. &SVM & $54.46$ & $74.76$ & $67.51$ &$84.67$ & $60.94$ & $79.93$ & $75.84$ & $89.26$\tabularnewline
\midrule 
Tran.&LR + ICI & $66.80$ & $79.26$ & $80.79$ & $87.92$ & $73.97$ & $84.13$ & $88.06$ & $92.53$\tabularnewline
Tran. &SVM + ICI & $65.77$ & $78.94$ &$80.56$  &$87.93$ & $73.16$ & $83.72$ & $87.87$ & $92.38$ \tabularnewline
\midrule 
Semi. &SVM + ICI ($15$/$15$) & $64.81$ & $78.11$ & $79.72$ &$87.39$  & $72.52$ &$83.23$  & $86.83$ &$91.58$ \tabularnewline
Semi. &SVM + ICI ($30$/$50$) & $68.24$ & $79.25$ & $83.14$ &$88.58$ & $75.50$ &$84.00$  & $88.94$  &$92.14$ \tabularnewline
Semi. &LR + ICI ($15$/$15$) & $65.86$ & $78.87$ & $81.10$ &$87.83$ & $73.67$ & $83.85$ & $87.28$ & $92.18$ \tabularnewline
Semi. &LR + ICI ($30$/$50$) & $69.66$ & $80.11$ & $84.01$ &$89.00$ &  $76.51$ & $84.32$ & $89.58$ &$92.48$ \tabularnewline
Semi. &LR + ICI ($80$/$80$) & $\textbf{71.41}$ & $\textbf{81.12}$ & $\textbf{85.44}$ &$\textbf{89.12}$ & $\textbf{78.07}$ & $\textbf{84.76}$ & $\textbf{91.11}$ &$\textbf{92.98}$ \tabularnewline
\bottomrule
\end{tabular*}
\vspace{0.01mm}
\caption{\label{fig:tfsl results} Test accuracies over $600$ episodes on several datasets.  
Results with $\left(\cdot\right)^{\textcolor{red}{1}}$ are reported in~\cite{DBLP:journals/corr/abs-1904-04232}, 
with $\left(\cdot\right)^{\textcolor{red}{2}}$ are reported in~\cite{sun2019learning}, 
with $\left(\cdot\right)^{\textcolor{red}{3}}$ are reported in~\cite{lee2019meta}.
$\left(\cdot\right)^{\textcolor{red}{4}}$ is our implementation with the official code of~\cite{liu2018learning}. 
Methods denoted by $\left(\cdot\right)^*$ denotes ResNet-18 with input size $224\times224$, while $\left(\cdot\right)^{\dag}$ denotes ResNet-18 with input size $84\times84$. 
Our method and other alternatives use ResNet-12 with input size $84\times84$.
\textbf{In.} and \textbf{Tran.} indicate inductive and transductive setting, respectively. 
\textbf{Semi.} denotes semi-supervised setting where $(\cdot/\cdot)$ shows the number of unlabeled data available in $1$-shot and $5$-shot experiments.
}
\end{table*}
  
\subsection{Semi-supervised few-shot learning}
\mypar{Settings.} 
In the inference process, the unlabeled data from the corresponding category pool is utilized to help FSL. 
In our experiments, we report following settings of SSFSL: 
(1) we use $15$ unlabeled samples for each class, the same as TFSL, to compare our algorithm in SSFSL and TFSL settings; 
(2) we use $30$ unlabeled samples in $1$-shot task, and $50$ unlabeled samples in $5$-shot task, the same as current SSFSL approaches~\cite{sun2019learning}; 
(3) we use $80$ unlabeled samples, to show the effectiveness of ICI compared with FSL algorithms with a larger network and higher-resolution inputs.
We denote these as ($15$/$15$), ($30$/$50$) and ($80$/$80$) in Table~\ref{fig:tfsl results}. 
Note that CUB is a fine-grained dataset and does not have so sufficient samples in each class, so we simply choose $5$ as support set, $15$ as query set and other samples as unlabeled set (about $39$ samples on average) in the $5$-shot task in the latter two settings. 
For all settings, we select $5$ samples for every class in each iteration. The process is finished when at most five instances for each class are excluded from the expanded support set. \ie, select ($10$/$10$), ($25$/$45$), ($75$/$75$) unlabeled instances in total. 
Further, we utilize Logistic Regression (denoted as \emph{LR}) and linear Support Vector Machine (denoted as \emph{SVM}) to show the robustness of ICI against different linear classifiers. 

\mypar{Competitors.} 
We compare our algorithm with current approaches in SSFSL. TPN~\cite{liu2018learning} uses labeled support set and unlabeled set to propagate label to one query sample each time. LST~\cite{sun2019learning} also uses self-taught learning strategy to pseudo-label data and select confident ones, but they do this by a neural network trained in the meta-learning manner for many iterations. 
Other approaches include Masked Soft k-Means~\cite{ren2018meta} and a combination of MTL with TPN and Masked Soft k-Means reported by LST.

\mypar{Results.} are shown in Table~\ref{fig:tfsl results} where denoted as Semi. in the first column. 
Analysis from the experimental results, we can find that:
(1) Compare SSFSL with TFSL with the same number of unlabeled data, we can see that our SSFSL results are only reduced by a little or even beat TFSL results, which indicates that the information we got from the unlabeled data are robust and we can indeed handle the true distribution with unlabeled data practically.
(2) The more unlabeled data we get, the better performance we have. Thus we can learn more knowledge with more unlabeled data almost consistently using a linear classifier (\eg logistic regression). When lots of unlabeled data are accessible, ICI achieves state-of-the-art in all experiments even compared with competitors which use bigger network and higher-resolution inputs. 
(3) Compared with other SSFSL approaches, ICI also achieves varying degrees of improvements in almost all tasks and datasets. These results further indicate the robustness of our algorithm. Compared logistic regression with SVM, the robustness of ICI still holds.

\subsection{Transductive few-shot learning}
\mypar{Settings.} 
In transductive few-shot learning setting, we have chance to access the query data in the inference stage. 
Thus the unlabeled set and the query dataset are the same. 
In our experiments, we select $5$ instances for each class in each iteration and repeat our algorithm until all the expected query samples are included, \ie, each class will be expanded by at most $15$ images.
We also utilize both Logistic Regression and SVM as our classifier, respectively.

\mypar{Competitors.}
We compare ICI with current TFSL approaches. 
TPN~\cite{liu2018learning} constructs a graph and uses label propagation to transfer label from support samples to query samples and learn their framework in a meta-learning way.
TEAM~\cite{qiao2019transductive} utilizes class prototypes with a data-dependent metric to inference labels of query samples.

\mypar{Results.} are shown in Table~\ref{fig:tfsl results} where denoted as Tran. in the first column. Experiments cross four benchmark datasets indicate that:
(1) Compared with basic linear classifier, ICI enjoys consistently improvements, especially in the 1-shot setting where the labeled data is extremely limited and  such improvements are robust regardless of utilizing which linear classifiers.
Further, compared results between \emph{mini}ImageNet and \emph{tiered}ImageNet, we can find that the improvement margin is in the similar scale, indicating that the improvement of ICI does not rely on the semantic relationship between base set and novel set.
Hence the effectiveness and robustness of ICI is confirmed practically. 
(2) Compared with current TFSL approaches, ICI also achieves the state-of-the-art results.
\subsection{Ablation study\label{subsec:Ablation-Study}}
\begin{figure}[h]
\begin{centering}
\includegraphics[width=1\columnwidth]{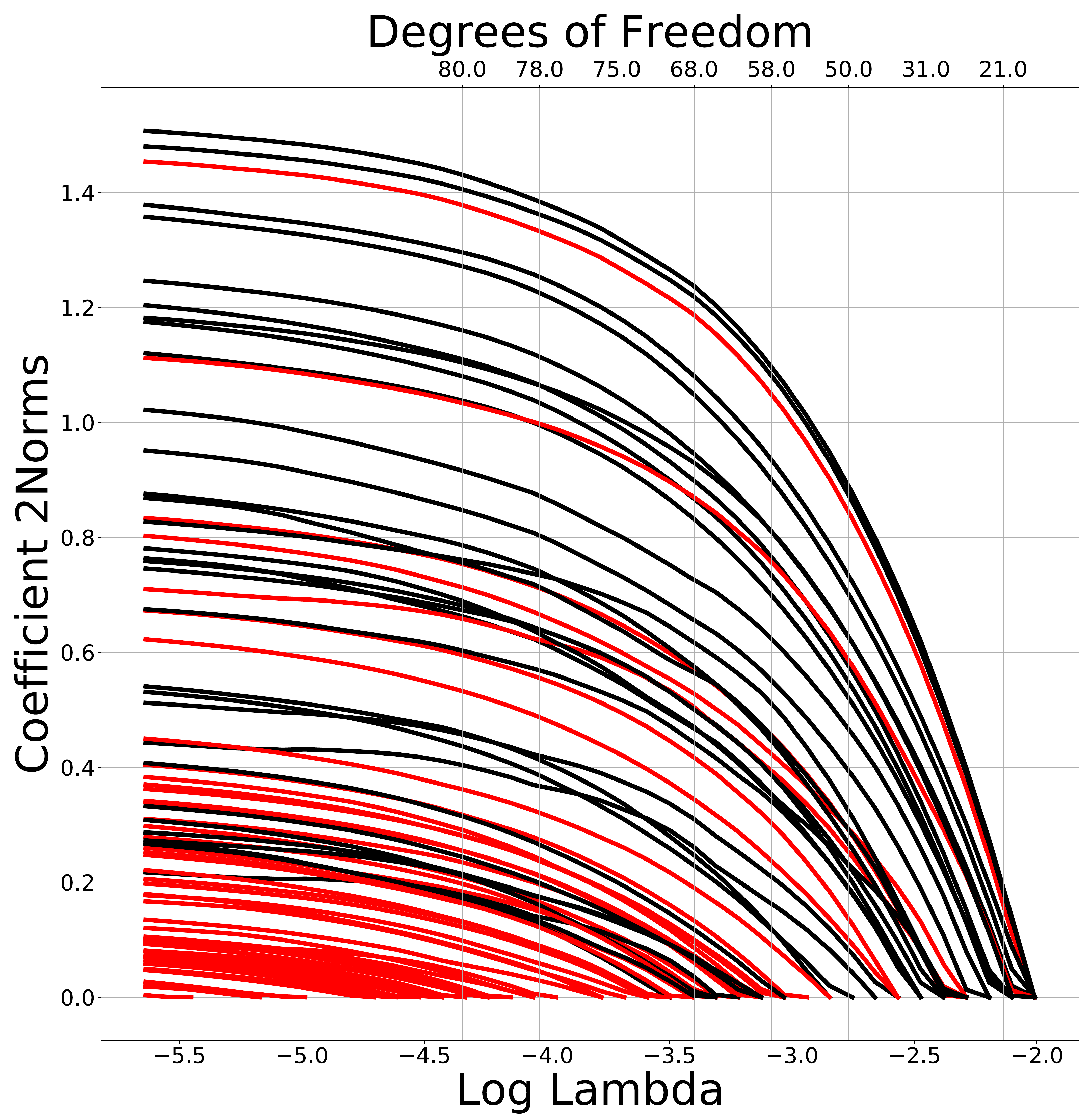}
\caption{\label{fig:effective}Regularization path of $\lambda$. Red lines are correct-predicted instances while black lines are wrong-predicted ones. ICI will choose instances in the lower-left subset.}
\end{centering}
\end{figure}
\mypar{Effectiveness of ICI.} To show the effectiveness of ICI, we visualize the regularization path of $\gamma$ in one episode of inference process in Figure~\ref{fig:effective} where red lines are instances that are correct-predicted while black lines are wrong-predicted ones. 
It is obvious that that most of the correct-predicted instances lie in the lower-left part. Since ICI will select samples whose norm will vanish in a lower $\lambda$. 
We could get more correct-predicted instances than wrong-predicted instances in a high ratio. 

\begin{table}[H]
\centering
\begin{tabular*}{\columnwidth}{@{\extracolsep{\fill}}lcccc}
\toprule 
\multirow{2}{*}{Model}&\multicolumn{2}{c}{Tran.}&\multicolumn{2}{c}{Semi.}\tabularnewline
&1shot&5shot&1shot&5shot\tabularnewline
\midrule
LR&$56.06$&$75.43$&$56.06$&$75.43$\tabularnewline
+ ra&$ 59.01$&$76.38$&$59.46 $&$76.58$\tabularnewline
+ nn&$63.24$&$77.63$&$63.10$&$77.75 $\tabularnewline
+ co&$63.29$&$77.92$&$63.57$&$77.71$\tabularnewline
 \midrule
ICI&   $\bf65.32$ & $\bf78.30$& $\bf64.60$& $\bf77.96$
\tabularnewline
\bottomrule
\end{tabular*}
\vspace{0.01mm}
\caption{\label{tab:ablation}
Compare to baselines on \emph{ mini}ImageNet under several settings.}
\end{table}
\mypar{Compare to baselines.} 
To further show the effectiveness of ICI, 
we compare ICI with other sample selection strategies under the self-taught learning pipeline. 
One simple strategy is randomly sampling the unlabeled data into the expanded support set in each iteration, denoted as \emph{ra}. 
Another is selecting the data based on the confidence given by the classifier, denoted by \emph{co}. 
In this strategy, 
the more confident the classifier is to one sample, 
the more trustworthy that sample is. 
The last one is replacing our algorithm of computing credibility by choosing the nearest-neighbor of each class in the feature space, denoted as \emph{nn}. 
In this part, we have $15$ unlabeled instances for each class and select $5$ to re-train the classifier by different methods for Semi. and Tran. task on \emph{ mini}ImageNet. 
From Table~\ref{tab:ablation},
we observe that ICI outperforms all the baselines in all settings.

\begin{figure}[h]
\centering
\includegraphics[width=1\columnwidth]{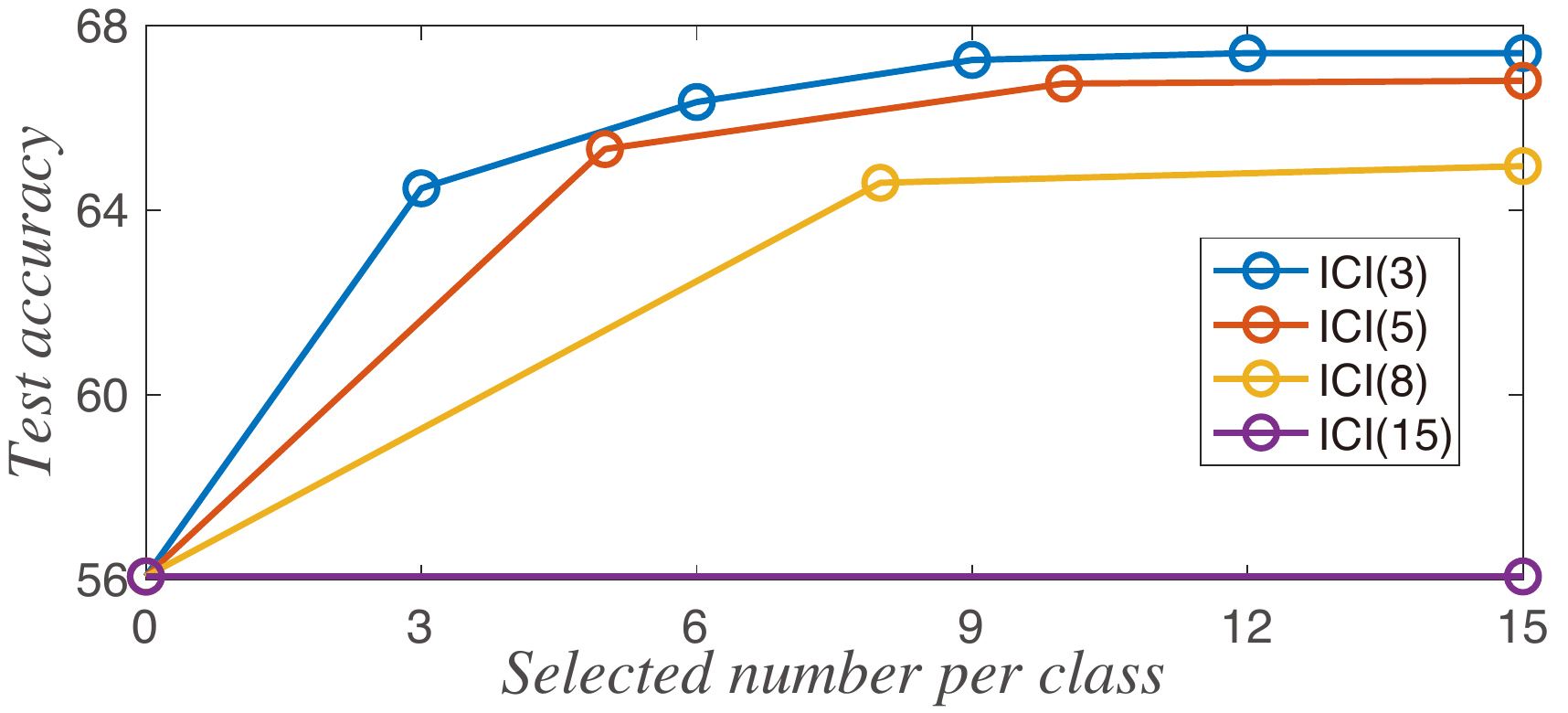} 
\caption{\label{fig:iter-manner}
Variation of accuracy as the selected samples increases over 600 episodes on \emph{mini}ImageNet. 
``ICI (\textit{n})'': select \textit{n} samples per class in each iteration.}
\end{figure}
\mypar{Effectiveness of iterative manner.}
Our intuition is the proposed ICI learns to generate a set of trustworthy unlabelled data for classifier training. 
Select
all the unlabelled data in \textit{one go}
cannot take the distribution, or the credibility of the unlabeled data into account, and thus produce more noise labels to hurt the performance of the model.
The classifier thus be trained with its prediction, resulting in no improvements in TFSL setting.
We briefly validate this as ICI (15) in Figure~\ref{fig:iter-manner} whilst ICI obtained better accuracy with iterative selection manner. 
For example, select $6$ images with two iterations (ICI(3)) is superior to select $8$ images in one iteration (ICI(8)).

\begin{table}[h]
\centering
\begin{tabular*}{\columnwidth}{@{\extracolsep{\fill}}lccccc}
\toprule 
Acc (\%)&0-10&10-20&20-30&30-40&40-50\tabularnewline
\midrule
b/t&0/0&0/0&1/3&16/23&105/125\tabularnewline
\midrule
\midrule
Acc (\%)&50-60&60-70&70-80&80-90&90-100\tabularnewline
\midrule
b/t&193/218&171/189&34/40&2/2&0/0\tabularnewline
\bottomrule
\end{tabular*}
\vspace{0.01mm}
\caption{\label{tab:stat}
We run 600 episodes, with each episode training an initial classifier.
We denote 
``Acc'' as the accuracy intervals; and 
``b/T'' as the number of classifiers experienced improvement v.s. 
total classifiers in this accuracy interval. }
\end{table}
\mypar{Robustness against initial classifier.} 
What are the requirements for the initial linear classifier? Is it necessary to satisfy that the accuracy of the initial linear classifier is higher than 50\% or even higher? 
The answer is no.
As long as the initial linear classifier can be trained, theoretically our method should work.  
It thus is a future open question of how the initial classifier affects. 
We briefly validate it in Table~\ref{tab:stat}.
We run 600 episodes, with each episode training an initial classifier with different classification accuracy.
Table~\ref{tab:stat}
shows that most classifiers can get improved by ICI regardless of the initial accuracy (even with accuracy of 30-40\%).

\mypar{Influence of reduced dimension.}  
In this part, we study the influence of reduced dimension $d$ in our algorithm on $5$-way $1$-shot \textit{mini}ImageNet experiments.
The results with reduced dimension $2$, $5$, $10$, $20$, $50$, and without dimensionality reduction \ie, $d=512$, are shown in Table~\ref{tab:reduced}. 
Our algorithm achieves better performance when the reduced dimension is much smaller than the number of instances (\ie, $d\ll n$), which is consistent with the theoretical property~\cite{fan2012partial}. 
Moreover, we can observe that our model achieves the best accuracy $66.80\%$ when $d=5$.
Practically, we adopt $d=5$ in our model.

\begin{table}[t!]
\begin{centering}
\begin{tabular*}{\columnwidth}{@{\extracolsep{\fill}}lclc}
\toprule
$d$ & Acc (\%)&Alg.&Acc (\%)\tabularnewline
\cmidrule{1-2} \cmidrule{3-4}
$2$ & $63.71\pm1.025$&Isomap~\cite{tenenbaum2000global} & $66.53\pm1.073$\tabularnewline
$5$ & $\bf66.80\pm1.096$&PCA~\cite{tipping1999probabilistic} & $66.80\pm1.096$\tabularnewline
$10$ & $66.25\pm1.048$&LTSA~\cite{zhang2004principal} & $64.61\pm1.058$\tabularnewline
$20$ & $64.98\pm1.049$&MDS~\cite{borg2003modern} & $59.99\pm0.941$\tabularnewline
$50$ & $61.54\pm0.980 $&LLE~\cite{roweis2000nonlinear} & $67.59\pm1.120 $\tabularnewline
$512$ & $57.41\pm0.877$&SE~\cite{belkin2003laplacian} & $67.70\pm1.117$ \tabularnewline 
\bottomrule
\end{tabular*}
\par\end{centering}
\vspace{2mm}
\caption{\label{tab:reduced}Influence of dimensionality reduction dimensions and algorithms.}
\end{table}

\mypar{Influence of dimension reduction algorithms.} 
Furthermore, we study the robustness of ICI to different dimension reduction algorithms.
We compare
Isomap~\cite{tenenbaum2000global},
principal components analysis~\cite{tipping1999probabilistic} (PCA),
local tangent space alignment~\cite{zhang2004principal} (LTSA),
multi-dimensional scaling~\cite{borg2003modern} (MDS),
locally linear embedding~\cite{roweis2000nonlinear} (LLE) and
spectral embedding~\cite{belkin2003laplacian} (SE)
on $5$-way $1$-shot \textit{mini}ImageNet experiments.
From Table~\ref{tab:reduced} we can observe that
ICI is robust across most of the dimensionality reduction algorithms (from LTAS $64.61\%$ to SE $67.7\%$) except MDS ($59.99\%$).
We adopt PCA for dimension reduction in our method.

\section{Conclusion}

In this paper, we have proposed a simple method, called Instance Credibility Inference (ICI) to exploit the distribution support of unlabeled instances for few-shot learning. 
The proposed ICI effectively select the most trustworthy pseudo-labeled instances according to their credibility to augment the training set. 
In order to measure the credibility of each pseudo-labeled instance, 
we propose to solve a linear regression hypothesis by increasing the sparsity of the incidental parameters~\cite{fan2012partial} and rank the pseudo-labeled instance with their sparsity degree. 
Extensive experiments show that our simple approach can establish new state-of-the-arts on four widely used few-shot learning benchmark datasets including \textit{mini}ImageNet, \textit{tiered}ImageNet, CIFAR-FS, and CUB.

\mypar{Acknowledgement.}
This work was supported in part by NSFC Projects (U1611461,61702108), 
Science and Technology Commission of Shanghai Municipality Projects (19511120700, 19ZR1471800), Shanghai Municipal Science and Technology Major Project (2018SHZDZX01), and  Shanghai Research and Innovation Functional Program (17DZ2260900).

{\small
\bibliographystyle{ieee_fullname}
\bibliography{main.bbl}

\begin{thebibliography}{10}\itemsep=-1pt

\bibitem{amini2002semi}
Massih-Reza Amini and Patrick Gallinari.
\newblock Semi-supervised logistic regression.
\newblock In {\em ECAI}, 2002.

\bibitem{arazo2019pseudo}
Eric Arazo, Diego Ortego, Paul Albert, Noel~E O'Connor, and Kevin McGuinness.
\newblock Pseudo-labeling and confirmation bias in deep semi-supervised
  learning.
\newblock {\em arXiv preprint arXiv:1908.02983}, 2019.

\bibitem{belkin2003laplacian}
Mikhail Belkin and Partha Niyogi.
\newblock Laplacian eigenmaps for dimensionality reduction and data
  representation.
\newblock {\em Neural computation}, 2003.

\bibitem{bennett1999semi}
Kristin~P Bennett and Ayhan Demiriz.
\newblock Semi-supervised support vector machines.
\newblock In {\em NeurIPS}, 1999.

\bibitem{bertinetto2018metalearning}
Luca Bertinetto, Joao~F. Henriques, Philip Torr, and Andrea Vedaldi.
\newblock Meta-learning with differentiable closed-form solvers.
\newblock In {\em ICLR}, 2019.

\bibitem{borg2003modern}
Ingwer Borg and Patrick Groenen.
\newblock Modern multidimensional scaling: Theory and applications.
\newblock {\em Journal of Educational Measurement}, 2003.

\bibitem{DBLP:journals/corr/abs-1904-04232}
Wei{-}Yu Chen, Yen{-}Cheng Liu, Zsolt Kira, Yu{-}Chiang~Frank Wang, and
  Jia{-}Bin Huang.
\newblock A closer look at few-shot classification.
\newblock In {\em ICLR}, 2019.

\bibitem{chen2019image}
Zitian Chen, Yanwei Fu, Yu-Xiong Wang, Lin Ma, Wei Liu, and Martial Hebert.
\newblock Image deformation meta-networks for one-shot learning.
\newblock In {\em CVPR}, 2019.

\bibitem{fan2012partial}
Jianqing Fan, Runlong Tang, and Xiaofeng Shi.
\newblock Partial consistency with sparse incidental parameters.
\newblock {\em arXiv preprint arXiv:1210.6950}, 2012.

\bibitem{finn2017model}
Chelsea Finn, Pieter Abbeel, and Sergey Levine.
\newblock Model-agnostic meta-learning for fast adaptation of deep networks.
\newblock In {\em ICML}, 2017.

\bibitem{ghiasi2018dropblock}
Golnaz Ghiasi, Tsung-Yi Lin, and Quoc~V Le.
\newblock Dropblock: A regularization method for convolutional networks.
\newblock In {\em NeurIPS}, 2018.

\bibitem{grandvalet2005semi}
Yves Grandvalet and Yoshua Bengio.
\newblock Semi-supervised learning by entropy minimization.
\newblock In {\em NeurIPS}, 2005.

\bibitem{DBLP:journals/corr/HeZRS15}
Kaiming He, Xiangyu Zhang, Shaoqing Ren, and Jian Sun.
\newblock Deep residual learning for image recognition.
\newblock {\em CoRR}, 2015.

\bibitem{he2016deep}
Kaiming He, Xiangyu Zhang, Shaoqing Ren, and Jian Sun.
\newblock Deep residual learning for image recognition.
\newblock In {\em CVPR}, 2016.

\bibitem{hilliard2018few}
Nathan Hilliard, Lawrence Phillips, Scott Howland, Art{\"e}m Yankov, Courtney~D
  Corley, and Nathan~O Hodas.
\newblock Few-shot learning with metric-agnostic conditional embeddings.
\newblock {\em arXiv preprint arXiv:1802.04376}, 2018.

\bibitem{huang2017densely}
Gao Huang, Zhuang Liu, Laurens Van Der~Maaten, and Kilian~Q Weinberger.
\newblock Densely connected convolutional networks.
\newblock In {\em CVPR}, 2017.

\bibitem{iscen2019label}
Ahmet Iscen, Giorgos Tolias, Yannis Avrithis, and Ondrej Chum.
\newblock Label propagation for deep semi-supervised learning.
\newblock In {\em CVPR}, 2019.

\bibitem{joachims1999transductive}
Thorsten Joachims.
\newblock Transductive inference for text classification using support vector
  machines.
\newblock In {\em ICML}, 1999.

\bibitem{krizhevsky2009learning}
Alex Krizhevsky, Geoffrey Hinton, et~al.
\newblock Learning multiple layers of features from tiny images.
\newblock Technical report, Citeseer, 2009.

\bibitem{krizhevsky2012imagenet}
Alex Krizhevsky, Ilya Sutskever, and Geoffrey~E Hinton.
\newblock Imagenet classification with deep convolutional neural networks.
\newblock In {\em NeurIPS}, 2012.

\bibitem{conf/iclr/LaineA17}
Samuli Laine and Timo Aila.
\newblock Temporal ensembling for semi-supervised learning.
\newblock In {\em ICLR}, 2017.

\bibitem{lee2013pseudo}
Dong-Hyun Lee.
\newblock Pseudo-label: The simple and efficient semi-supervised learning
  method for deep neural networks.
\newblock In {\em ICML workshops}, 2013.

\bibitem{lee2019meta}
Kwonjoon Lee, Subhransu Maji, Avinash Ravichandran, and Stefano Soatto.
\newblock Meta-learning with differentiable convex optimization.
\newblock In {\em CVPR}, 2019.

\bibitem{lemke2015metalearning}
Christiane Lemke, Marcin Budka, and Bogdan Gabrys.
\newblock Metalearning: a survey of trends and technologies.
\newblock {\em Artificial intelligence review}, 2015.

\bibitem{li2019finding}
Hongyang Li, David Eigen, Samuel Dodge, Matthew Zeiler, and Xiaogang Wang.
\newblock Finding task-relevant features for few-shot learning by category
  traversal.
\newblock In {\em CVPR}, 2019.

\bibitem{li2014towards}
Yu-Feng Li and Zhi-Hua Zhou.
\newblock Towards making unlabeled data never hurt.
\newblock {\em TPAMI}, 2014.

\bibitem{li2017meta}
Zhenguo Li, Fengwei Zhou, Fei Chen, and Hang Li.
\newblock Meta-sgd: Learning to learn quickly for few-shot learning.
\newblock {\em arXiv preprint arXiv:1707.09835}, 2017.

\bibitem{liu2018learning}
Yanbin Liu, Juho Lee, Minseop Park, Saehoon Kim, Eunho Yang, Sung~Ju Hwang, and
  Yi Yang.
\newblock Learning to propagate labels: Transductive propagation network for
  few-shot learning.
\newblock {\em arXiv preprint arXiv:1805.10002}, 2018.

\bibitem{miayto2016virtual}
Takeru Miayto, Andrew~M Dai, and Ian Goodfellow.
\newblock Virtual adversarial training for semi-supervised text classification.
\newblock 2016.

\bibitem{mishra2017simple}
Nikhil Mishra, Mostafa Rohaninejad, Xi Chen, and Pieter Abbeel.
\newblock A simple neural attentive meta-learner.
\newblock {\em arXiv preprint arXiv:1707.03141}, 2017.

\bibitem{munkhdalai2017rapid}
Tsendsuren Munkhdalai, Xingdi Yuan, Soroush Mehri, and Adam Trischler.
\newblock Rapid adaptation with conditionally shifted neurons.
\newblock {\em arXiv preprint arXiv:1712.09926}, 2017.

\bibitem{nichol2018first}
Alex Nichol, Joshua Achiam, and John Schulman.
\newblock On first-order meta-learning algorithms.
\newblock {\em arXiv preprint arXiv:1803.02999}, 2018.

\bibitem{oreshkin2018tadam}
Boris Oreshkin, Pau~Rodr{\'\i}guez L{\'o}pez, and Alexandre Lacoste.
\newblock Tadam: Task dependent adaptive metric for improved few-shot learning.
\newblock In {\em NeurIPS}, 2018.

\bibitem{qiao2019transductive}
Limeng Qiao, Yemin Shi, Jia Li, Yaowei Wang, Tiejun Huang, and Yonghong Tian.
\newblock Transductive episodic-wise adaptive metric for few-shot learning.
\newblock In {\em ICCV}, 2019.

\bibitem{self-taught-learning}
Rajat Raina, Alexis Battle, Honglak Lee, Benjamin Packer, and Andrew~Y. Ng.
\newblock Self-taught learning: Transfer learning from unlabeled data.
\newblock In {\em ICML}, 2007.

\bibitem{ravi2016optimization}
Sachin Ravi and Hugo Larochelle.
\newblock Optimization as a model for few-shot learning.
\newblock In {\em ICLR}, 2017.

\bibitem{ren2018meta}
Mengye Ren, Eleni Triantafillou, Sachin Ravi, Jake Snell, Kevin Swersky,
  Joshua~B Tenenbaum, Hugo Larochelle, and Richard~S Zemel.
\newblock Meta-learning for semi-supervised few-shot classification.
\newblock {\em arXiv preprint arXiv:1803.00676}, 2018.

\bibitem{roweis2000nonlinear}
Sam~T Roweis and Lawrence~K Saul.
\newblock Nonlinear dimensionality reduction by locally linear embedding.
\newblock {\em science}, 2000.

\bibitem{rusu2018meta}
Andrei~A Rusu, Dushyant Rao, Jakub Sygnowski, Oriol Vinyals, Razvan Pascanu,
  Simon Osindero, and Raia Hadsell.
\newblock Meta-learning with latent embedding optimization.
\newblock {\em arXiv preprint arXiv:1807.05960}, 2018.

\bibitem{shi2018transductive}
Weiwei Shi, Yihong Gong, Chris Ding, Zhiheng MaXiaoyu~Tao, and Nanning Zheng.
\newblock Transductive semi-supervised deep learning using min-max features.
\newblock In {\em ECCV}, 2018.

\bibitem{simon2013blockwise}
Noah Simon, Jerome Friedman, and Trevor Hastie.
\newblock A blockwise descent algorithm for group-penalized multiresponse and
  multinomial regression.
\newblock {\em arXiv preprint arXiv:1311.6529}, 2013.

\bibitem{simonyan2014very}
Karen Simonyan and Andrew Zisserman.
\newblock Very deep convolutional networks for large-scale image recognition.
\newblock {\em arXiv preprint arXiv:1409.1556}, 2014.

\bibitem{snell2017prototypical}
Jake Snell, Kevin Swersky, and Richard Zemel.
\newblock Prototypical networks for few-shot learning.
\newblock In {\em NeurIPS}, 2017.

\bibitem{JMLR:v15:srivastava14a}
Nitish Srivastava, Geoffrey Hinton, Alex Krizhevsky, Ilya Sutskever, and Ruslan
  Salakhutdinov.
\newblock Dropout: A simple way to prevent neural networks from overfitting.
\newblock {\em JMLR}, 2014.

\bibitem{sun2019learning}
Qianru Sun, Xinzhe Li, Yaoyao Liu, Shibao Zheng, Tat-Seng Chua, and Bernt
  Schiele.
\newblock Learning to self-train for semi-supervised few-shot classification.
\newblock {\em arXiv preprint arXiv:1906.00562}, 2019.

\bibitem{sung2018learning}
Flood Sung, Yongxin Yang, Li Zhang, Tao Xiang, Philip~HS Torr, and Timothy~M
  Hospedales.
\newblock Learning to compare: Relation network for few-shot learning.
\newblock In {\em CVPR}, 2018.

\bibitem{sung2017learning}
Flood Sung, Li Zhang, Tao Xiang, Timothy Hospedales, and Yongxin Yang.
\newblock Learning to learn: Meta-critic networks for sample efficient
  learning.
\newblock {\em arXiv preprint arXiv:1706.09529}, 2017.

\bibitem{tarvainen2017mean}
Antti Tarvainen and Harri Valpola.
\newblock Mean teachers are better role models: Weight-averaged consistency
  targets improve semi-supervised deep learning results.
\newblock In {\em NeurIPS}, 2017.

\bibitem{tenenbaum2000global}
Joshua~B Tenenbaum, Vin De~Silva, and John~C Langford.
\newblock A global geometric framework for nonlinear dimensionality reduction.
\newblock {\em science}, 2000.

\bibitem{tipping1999probabilistic}
Michael~E Tipping and Christopher~M Bishop.
\newblock Probabilistic principal component analysis.
\newblock {\em Journal of the Royal Statistical Society: Series B (Statistical
  Methodology)}, 1999.

\bibitem{triantafillou2017few}
Eleni Triantafillou, Richard Zemel, and Raquel Urtasun.
\newblock Few-shot learning through an information retrieval lens.
\newblock In {\em NeurIPS}, 2017.

\bibitem{vapnik1998statistical}
Vladimir Vapnik and Vlamimir Vapnik.
\newblock Statistical learning theory wiley.
\newblock {\em New York}, 1998.

\bibitem{vinyals2016matching}
Oriol Vinyals, Charles Blundell, Timothy Lillicrap, Daan Wierstra, et~al.
\newblock Matching networks for one shot learning.
\newblock In {\em NeurIPS}, 2016.

\bibitem{wah2011caltech}
C. Wah, S. Branson, P. Welinder, P. Perona, and S. Belongie.
\newblock {The Caltech-UCSD Birds-200-2011 Dataset}.
\newblock Technical report, California Institute of Technology, 2011.

\bibitem{NoisyStudent}
Qizhe Xie, Eduard Hovy, Minh-Thang Luong, and Quoc~V. Le.
\newblock Self-training with noisy student improves imagenet classification.
\newblock {\em arXiv preprint arXiv:1911.04252}, 2019.

\bibitem{yoon2019tapnet}
Sung~Whan Yoon, Jun Seo, and Jaekyun Moon.
\newblock Tapnet: Neural network augmented with task-adaptive projection for
  few-shot learning.
\newblock {\em arXiv preprint arXiv:1905.06549}, 2019.

\bibitem{yosinski2014transferable}
Jason Yosinski, Jeff Clune, Yoshua Bengio, and Hod Lipson.
\newblock How transferable are features in deep neural networks?
\newblock In {\em NeurIPS}, 2014.

\bibitem{zhang2017learning}
Li Zhang, Tao Xiang, and Shaogang Gong.
\newblock Learning a deep embedding model for zero-shot learning.
\newblock In {\em CVPR}, 2017.

\bibitem{zhang2004principal}
Zhenyue Zhang and Hongyuan Zha.
\newblock Principal manifolds and nonlinear dimensionality reduction via
  tangent space alignment.
\newblock {\em SIAM journal on scientific computing}, 2004.

\end{thebibliography}
}
\end{document}